\documentclass[letterpaper]{article} % DO NOT CHANGE THIS
\usepackage{aaai25}  % DO NOT CHANGE THIS
\usepackage{times}  % DO NOT CHANGE THIS
\usepackage{helvet}  % DO NOT CHANGE THIS
\usepackage{courier}  % DO NOT CHANGE THIS
\usepackage[hyphens]{url}  % DO NOT CHANGE THIS
\usepackage{graphicx} % DO NOT CHANGE THIS
\usepackage{natbib}  % DO NOT CHANGE THIS AND DO NOT ADD ANY OPTIONS TO IT
\usepackage{caption} % DO NOT CHANGE THIS AND DO NOT ADD ANY OPTIONS TO IT
\usepackage{algorithm}
\usepackage{algorithmic}

\usepackage{makecell}

\title{Reference-Based Post-OCR Processing with LLM for \\Precise Diacritic Text in Historical Document Recognition}
\author{
    Thao Do,
    Dinh Phu Tran,
    An Vo,
    Daeyoung Kim
}
\affiliations{
    Korea Advanced Institute of Science and Technology (KAIST), Daejeon 34141, South Korea\\
    \{thaodo, phutx2000, an.vo, kimd\}@kaist.ac.kr
}

\usepackage{bibentry}
\begin{document}

\maketitle

\begin{abstract}
Extracting fine-grained OCR text from aged documents in diacritic languages remains challenging due to unexpected artifacts, time-induced degradation, and lack of datasets. While standalone spell correction approaches have been proposed, they show limited performance for historical documents due to numerous possible OCR error combinations and differences between modern and classical corpus distributions.
We propose a method utilizing available content-focused ebooks as a reference base to correct imperfect OCR-generated text, supported by large language models. This technique generates high-precision pseudo-page-to-page labels for diacritic languages, where small strokes pose significant challenges in historical conditions. The pipeline eliminates various types of noise from aged documents and addresses issues such as missing characters, words, and disordered sequences.
Our post-processing method, which generated a large OCR dataset of classical Vietnamese books, achieved a mean grading score of 8.72 on a 10-point scale. This outperformed the state-of-the-art transformer-based Vietnamese spell correction model, which scored 7.03 when evaluated on a sampled subset of the dataset. We also trained a baseline OCR model to assess and compare it with well-known engines. Experimental results demonstrate the strength of our baseline model compared to widely used open-source solutions. The resulting dataset will be released publicly to support future studies.
\end{abstract}

% Uncomment the following to link to your code, datasets, an extended version or similar.
\begin{links}
    \link{Dataset}{https://github.com/thaodod/VieBookRead}
\end{links}

\section{Introduction} \label{introduction}

Optical character recognition (OCR) utilizing deep learning methodologies for pervasive languages, such as English, has attained great performance for digitizing book-form documents, and the research community focus has thus transitioned to new challenges. Conversely, for languages replete with diacritics, such as Vietnamese (67 diacritic letters), Slavic (25 letters), Czech (19 letters) \cite{stankevivcius2022correcting}, hopes of similar performance remain unrealized. The problem becomes challenging when these tiny strokes of diacritics faded in historical documents like books or newspapers, and several background problems emerge such as ripped, taped or obscured words, smeared inks, watermarks, third party seals, termite bites and bleed-through.
Digitization and preservation of historical texts create substantial social impact, especially as social science researchers and heritage offices struggle to reliably collect and retrieve the data needed for their work, research, and education activities. % just added this after review %

Several projects \cite{gooding2013myth} digitize these aged books by capturing page images to preserve them electronically. The next step is to transcribe images for text-based storage and later text mining, which is also a time-consuming process. Many earlier industrial tools and modern AI models were developed for this purpose. However, the precision and quality of the text generated automatically remain inadequate for uncommon languages. So, proofreading and manual modifications are necessary following these processes. To alleviate this issue, researchers leveraged the well-known task of spelling check from office text editing to correct post-OCR texts similarly. Some are based on statistical approaches to offer a list of candidates for replacing words, while recent methods offer full-text (e.g., paragraph) transformation based on the context and distribution the model was trained on beforehand \cite{hladek2020survey}.

Nevertheless, the kind of self-sufficient spell correction model has some limitations. Let's take the letter `a' in Vietnamese, it has 17 variants with diacritics, the same is true for their `o'. So, ``con ga'' could have $18\times18$ combinations (assuming consonants are correct) and 10 of them are meaningful. Therefore, one limitation of the standalone spell correction model is that it heavily depends on OCR capability because the more errors are made (vowels, consonants, diacritics), the more combinations and candidates are created \cite{hladek2020survey}. This leads to a lesser chance of generating correct text. The second problem especially comes with seq2seq style models, which were trained on recent corpus. Those trained sets have a different distribution (modern contents) compared with the distribution of classical contents where archaic, obsolete vocabulary and old-fashioned grammar exist. Some tried to directly train from suitable distribution but with the cost of annotation data.

By chance, we discovered that, before recent advancements in AI, from the 1990s, readers and preservation enthusiasts made efforts to digitize aged books in electronic form. Ebook makers usually focus on the main content of the book as it is the final form fitting on ebook readers which provides flexibility to the screen size where pagination is less important. Thus, intermediate processed records (i.e. page transcriptions) were either discarded or stored with less diligence on local storage and lost over time due to hardware failures and the absence of widely adopted personal cloud storage until around 2012.

This situation occurs within a Vietnamese forum\footnote{\url{https://tve-4u.org/}} where book enthusiasts collaborate to transcribe, proofread, and produce ebook files, without page-separated transcriptions. It is noteworthy that Vietnamese classical document records were published during a complex transitional period in history, leading to non-standardized grammar and orthography compared to contemporary norms. Consequently, similar yet distinct words may be used synonymously by different authors from various regions, besides, texts contain archaic, obsolete lexis. Fortunately, participants in the preservation attempt to preserve the original orthography of the texts. These ebook files are hypothesized to be valuable reference sources for document digitization and later downstream tasks. Similar sites also exist in other languages such as French e.g., project Gutenberg\footnote{\url{https://www.gutenberg.org/ebooks/bookshelf/313}}.

To achieve decent accuracy, a supervised OCR method is preferred; however, we cannot directly leverage the content-focus ebooks for training, as they are not page-to-page (page2page) mapping. To solve the issue, we propose a method with the goal of using these ebooks to eventually obtain a reliable page2page label. In general, we first reduce noises by a few heuristic filters. Next, consider content-focus ebooks as \textit{reference sources}, imperfect OCR-generated text can then be corrected by a large language model (LLM) based on semantic and visual similarity between them and the \textit{reference}. The final pseudo label is returned after short-length, dangling texts are adjusted by LLM-based spelling correction with same-page context. Details of the full pipeline are described in Section \ref{method}.

To verify our hypothesis, we gathered a collection of rare, aged books published from 1850 onward, along with corresponding ebooks. This effort also served to generate a high-precision benchmark dataset for Vietnamese, as a public dense-text dataset of books with historical characteristics is still not available. Based on evaluations of the dataset, our post-processing yields better quality labels with a human-based evaluation rate of $8.72/10$ (and 86.66\% semantic better), compared to only $7.03/10$ (and 6.81\% semantic better) of the state-of-the-art spell correction approach. We also trained 2 baselines using our final pseudo labels to compare with contemporary models like GPT-4o, Google Document AI, and Tesseract. Results at this point showed that GPT-4o achieved the top metrics, while our baseline outperformed known open-source solutions.

To summarize, we proposed a novel method to utilize available ebook resources as a reference base to correct post-OCR text. The method employs LLMs to generate fine-grained pseudo ground truth without additional data annotation costs while maintaining the historical characteristics of classical documents. The methodology is applicable not only to Vietnamese but also to other diacritical languages, creating a basis to generate a quality page2page dataset to train a better open-source OCR model for underserved communities. A second main contribution is, that we introduce a large Vietnamese publicly available dataset named as VieBookRead for classical books from the 19th century, which possess unique characteristics that are uncommon in contemporary works to close the gap in existing datasets and facilitate later research.

\section{Related Work}
\subsection{Vietnamese Document and Textual Image Datasets}
Related to the preservation of historical script records, \cite{dang2022nomnaocr} collected the largest handwritten pages (NomNaOCR) from the Nom Preservation Foundation that were written in Nom script. This former writing system is the logographic script used to write Vietnamese between the 15th and early 20th centuries. On the other hand, \cite{van2024vitextvqa} tackles a recent task of Visual Question Answering in current Latin-based Vietnamese by introducing the ViTextVQA set which composes scene text images (e.g., billboards, signs, banners). Meanwhile, \cite{le2023dataset} organized a limited set named VNDoC, which contains legal, administrative scanned documents from companies, and the government (public availability could not be verified); \cite{thi2022novel} introduced a dataset (Vi-BCI) mainly for extracting information from book's cover images.

A significant gap exists in Vietnamese historical research due to the lack of datasets for valuable books from the 19th century through the early 2000s. This period marked the birth of modern Vietnamese script and saw a surge in book publishing, driven by printing machinery. Also by efforts of intellectuals worked to transfer earlier historical records and knowledge to the modern script, aiming to improve education under the French colony. Publications from this era possess unique characteristics that are uncommon in contemporary works, reflecting the transitional nature of the period. Thus, a Vietnamese document dataset for this era is needed to facilitate further historical studies.

\subsection{Spelling, Diacritics, and Typos Correction}
Regarding the post-processing after the OCR and document analysis step, spelling (typos) correction or diacritics restoration techniques are sometimes leveraged to improve text recognition results in many diacritic languages. \cite{ruch2003using} proposed lexical disambiguation and named-entity recognition to correct spelling in the patient record in French; while \cite{atserias2012spell} used lexical co-occurrence to decide which word variants for a given context to perform spell check in Spanish. Recently, \cite{stankevivcius2022correcting} utilized a ByT5 to perform diacritics restoration on benchmark datasets of 12 languages such as Croatian, Latvian with high word-level accuracy. 

\begin{figure}[ht]
    \centering
    \includegraphics[width=0.474\textwidth]{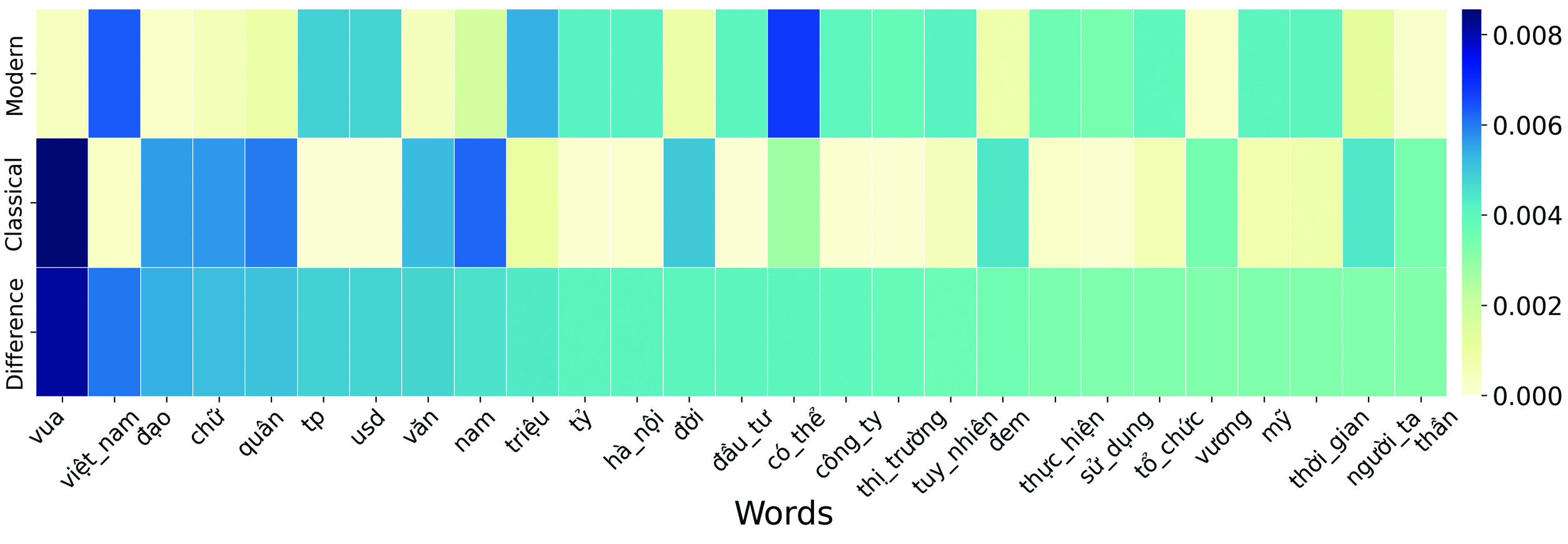}
    \caption{TF-IDF differences between a modern corpus vs our classic set book content}
    \label{fig:tfidf}
\end{figure}

Similarly but specifically for Vietnamese, \cite{nguyen2008vietnamese} used bi-gram with other heuristics to pick a good candidate in the task spell-check and correction in online news texts. Embrace deep learning prevalent in the seq2seq model around 2019 \cite{nguyen2020deep} introduced a seq2seq model based on LSTM architecture to adjust input sentences in a manner similar to text translation. Recently, \cite{do2021vsec, tran2021hierarchical} from a popular Vietnamese social network company proposed a similar approach, but using an advanced transformer backbone with a larger training corpus to achieve competitive performances. Still, standalone spell correction models have certain drawbacks: The accuracy of the output significantly relies on the quality of the OCR, as an increased number of errors (vowels, consonants, diacritics, punctuation) generates more combinations with lower confidence, reducing the likelihood of producing correct text. Another issue particularly relevant to seq2seq models trained on recent corpora applying to classical settings is the distributional difference in the train set. We highlight the difference in word TF-IDF between the modern corpus used by \cite{do2021vsec} and the corpus from our ebook texts in Figure \ref{fig:tfidf}.

Most recently, \cite{thomas2024leveraging} taught Llama 2 \cite{touvron2023llama} to correct the post-OCR text of historical newspapers in English in the same manner (i.e., seq2seq transformation). The method proved the potential but came with a requirement to prepare sequence pairs (i.e., post-OCR faulty text and its ground truth). And, hallucination could be fairly high under no constraint on context or reference, especially when the component words in the text are visually similar but semantically different like in diacritic languages.

\section{Dataset Preparation \& Analysis} \label{data}

\subsection{Data Collection}
Due to the lack of available datasets for classic Vietnamese books as we mentioned, and to test our proposed methodology for post-OCR processing of a diacritic language, we created a dataset called VieBookRead.
Initially, we followed threads on ebook forums with verified accounts, where users had listed all books that had been completed by proofreading for preservation. We then downloaded the available ebooks (e.g. epub format) and scraped including searched, verified, and fetched the corresponding scanned PDF files from the Internet, as the forum did not provide the scanned versions.

\subsection{Dataset Analysis}

VieBookRead dataset focuses on aged documents with historical characteristics. A new high-quality dataset is crucial for improving text extraction from images. The dataset is diverse, covering various fields. Some statistics about the VieBookRead dataset are as follows:

\begin{itemize}
    \item The dataset includes 123 books with epubs and 50 books without epubs published between 1850-2003.
    \item 123 books having ebooks extracted to 27,025 pages, corresponding to 27,025 images with resolution from $472 \times 719$ up to $6093 \times 9000$. 25,585 pages are finally kept after clipping off book covers and other irrelevant pages (i.e. some first and last pages)
    \item The content within the books includes 4 languages with approximately 87.20\% Vietnamese, followed by 7.62\% English, 5.16\% French, and 0.02\% Chinese.
    \item We divide the dataset into 10 major genres: history, biography, geography, literature, education,  novel, culture, philosophy, folk, and sociology. Some books can be classified into multiple categories.
\end{itemize}

\subsection{Comparison with Other Datasets}

\begin{figure}[ht]
    \centering
    \includegraphics[width=0.4\textwidth]{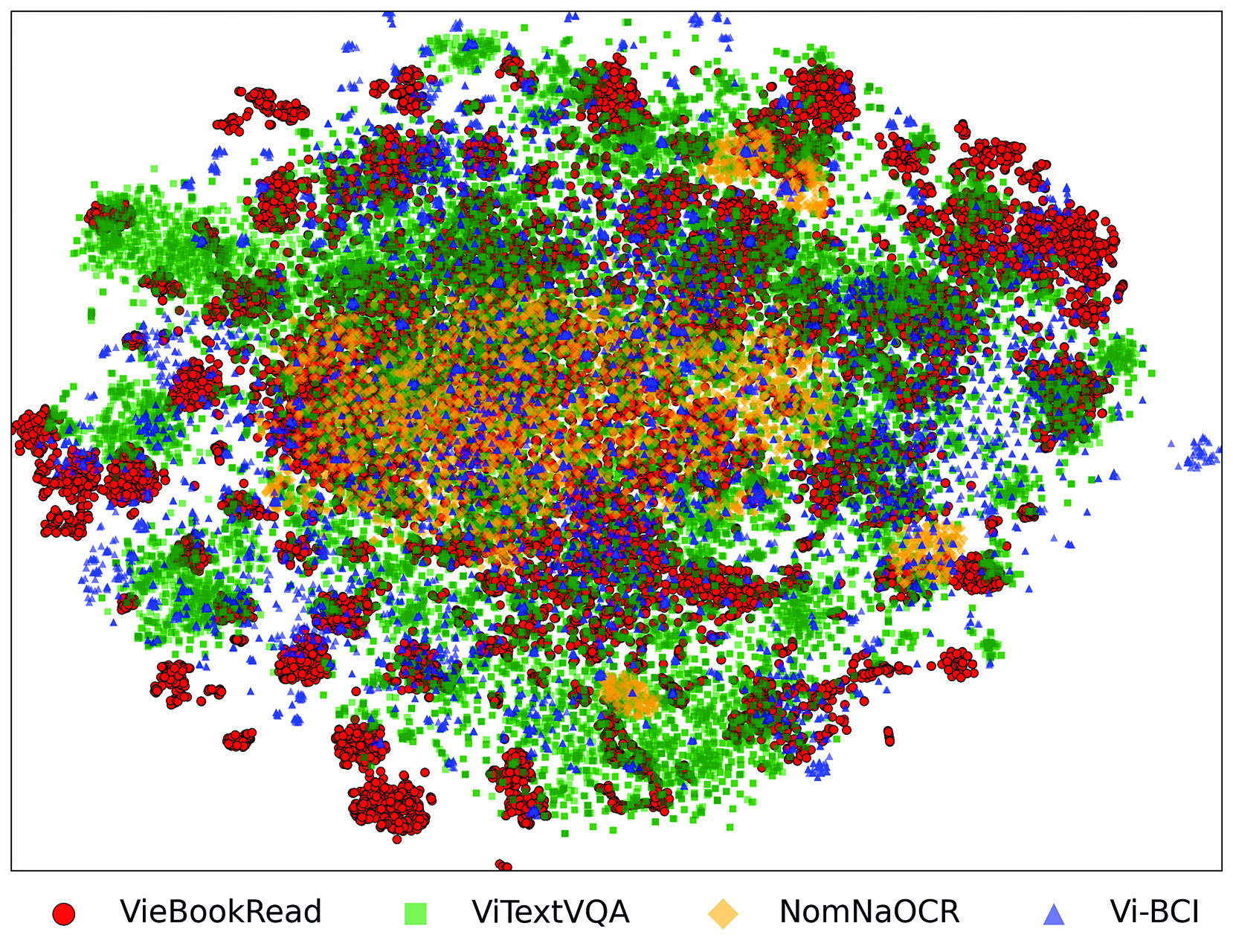}
    \caption{t-SNE visualization for Ours vs ViTextVQA, NomNaOCR, Vi-BCI using features by CLIP ViT-L/14@336px image encoder}
    \label{fig:image2}
\end{figure}

To illustrate the correlation and coverage of our dataset alongside other datasets from a big-picture view, we use the OpenAI CLIP \cite{radford2021learning} ViT-L/14@336px model to extract features for visualization although CLIP was not mainly trained by document samples. We compare ours with other recent document datasets in Vietnamese, including ViTextVQA, NomNaOCR, and Vi-BCI in the latent space as Fig \ref{fig:image2}.

Our dataset in red points shows a number of areas of coverage beyond what has been achieved in previous research. Additionally, Table \ref{tab:data-compare} compares some aspects of datasets: our VieBookRead has a comparable large size of 25,585 pages of book images written mainly in Vietnamese but also in other 3 languages occasionally. The features pose a uniqueness compared with other datasets which are scene texts, Nom scripts, and book covers already covered in the previous works. Few samples are included in the Appendix.

\begin{table}[ht]
    \setlength{\abovecaptionskip}{2pt}
    \setlength{\belowcaptionskip}{0pt}
    \centering
    \resizebox{\columnwidth}{!}{%
        \begin{tabular}{llll}
            \textbf{Dataset} & \textbf{\makecell[l]{\# of\\images}} & \textbf{Language} & \textbf{Main content} \\ \hline
            ViTextVQA & 16,762 & Vietnamese & \makecell[l]{Scene text boards,\\signs, posters} \\ \hline
            NomNaOCR & 2,953 & \makecell[l]{Logographic Vietnamese\\(Former script)} & Ancient books \\ \hline
            Vi-BCI & 7,875 & Vietnamese & Book covers \\ \hline
            VieBookRead & 25,585 & \makecell[l]{Vietnamese, English,\\French, Chinese} & Classical books \\ \hline
        \end{tabular}
    }
    \caption{Related datasets comparison}
    \label{tab:data-compare}
\end{table}

\section{Methodology} \label{method}
\begin{figure*}[ht]
    \centering
    \includegraphics[width=0.85\textwidth]{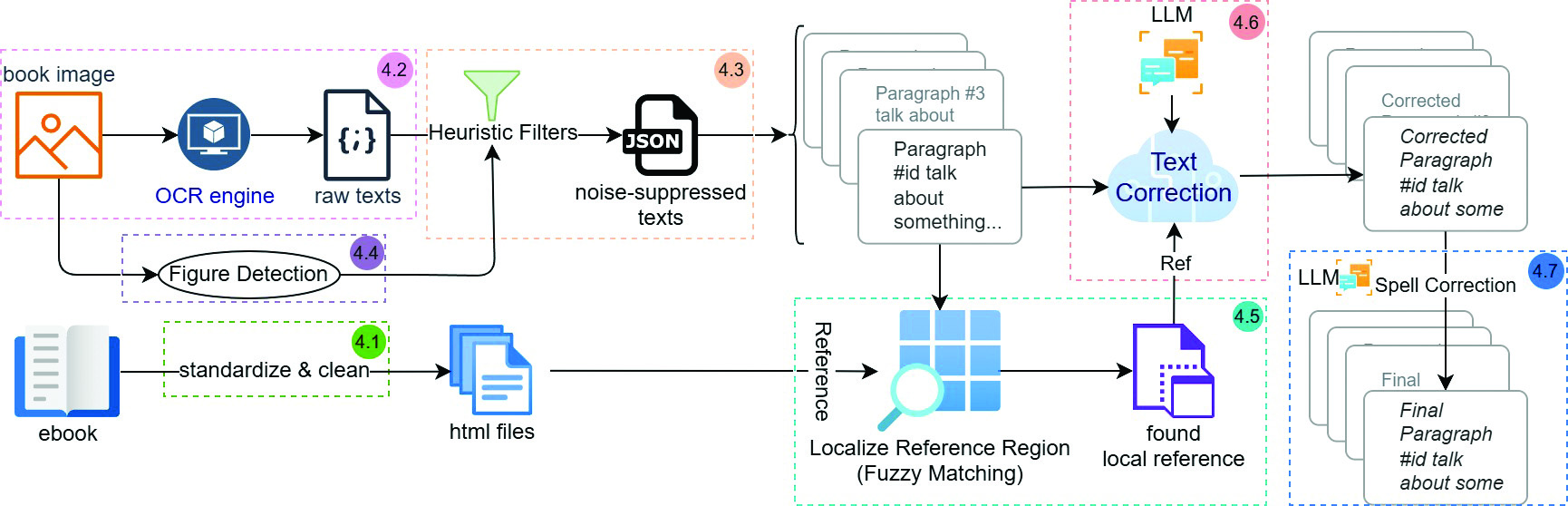}
    \caption{Our main pipeline to generate precise diacritic text for book images}
    \label{fig:pipeline}
\end{figure*}
Our overall pipeline to obtain reliable pseudo page2page labels is illustrated in Figure \ref{fig:pipeline}. The specifics of each step are discussed in the following subsections.

\subsection{Data Pre-processing}
On the side of images (i.e. captured images of pages), we verify content to avoid insertions of irrelevant pages from image providers, and also separate pages from multi-page scan setups in some cases. Note that because there are several cases where the same book name has different versions by different publishers, we make sure to fetch corresponding scanned files for the release ebook file to ensure the quality of later matching steps.

On the other hand, ebook files are standardized to epub files and then extracted all to HTML (.xml) files where we can clean up all redundant recursive tags and unnecessary tags that don't provide useful information but also cause noise for later processing. Due to the inconsistency of software development in character encoding in the early 2000s, countable files having visually identical text but indeed encoded by different character sets thus caused confusion in fuzzy matching. This is an example of why we have to standardize by re-encoding to the same character set in text processing to avoid various errors later.

\subsection{Extract texts by OCR model}
From empirical observation, we chose Azure Document Intelligence (ADI) as the main OCR engine for page reading because of its high recognition quality and holistic information. Results provide where to have word-level, line-level and paragraph-level extraction with its polygon and bounding boxes. We learned that paragraph level is inconsistently reliable because of the high degree of noise as explained in Section \ref{introduction}. Thus, we leverage line-level elements, where they are more reliable, to eliminate a part of noises, but still not aggressively to preserve actual texts of the content. The detail of how we reduce noise is described in the next subsection.

We believe that the selection and performance of the OCR model to extract text in the first place lead to a different effect on the final results of our pipeline. However, our correction algorithm in the later part remains significantly useful for improving the final text even for those outcomes from the open source OCR model such as Tesseract.

\begin{figure}[ht]
    \centering
    \includegraphics[width=0.474\textwidth]{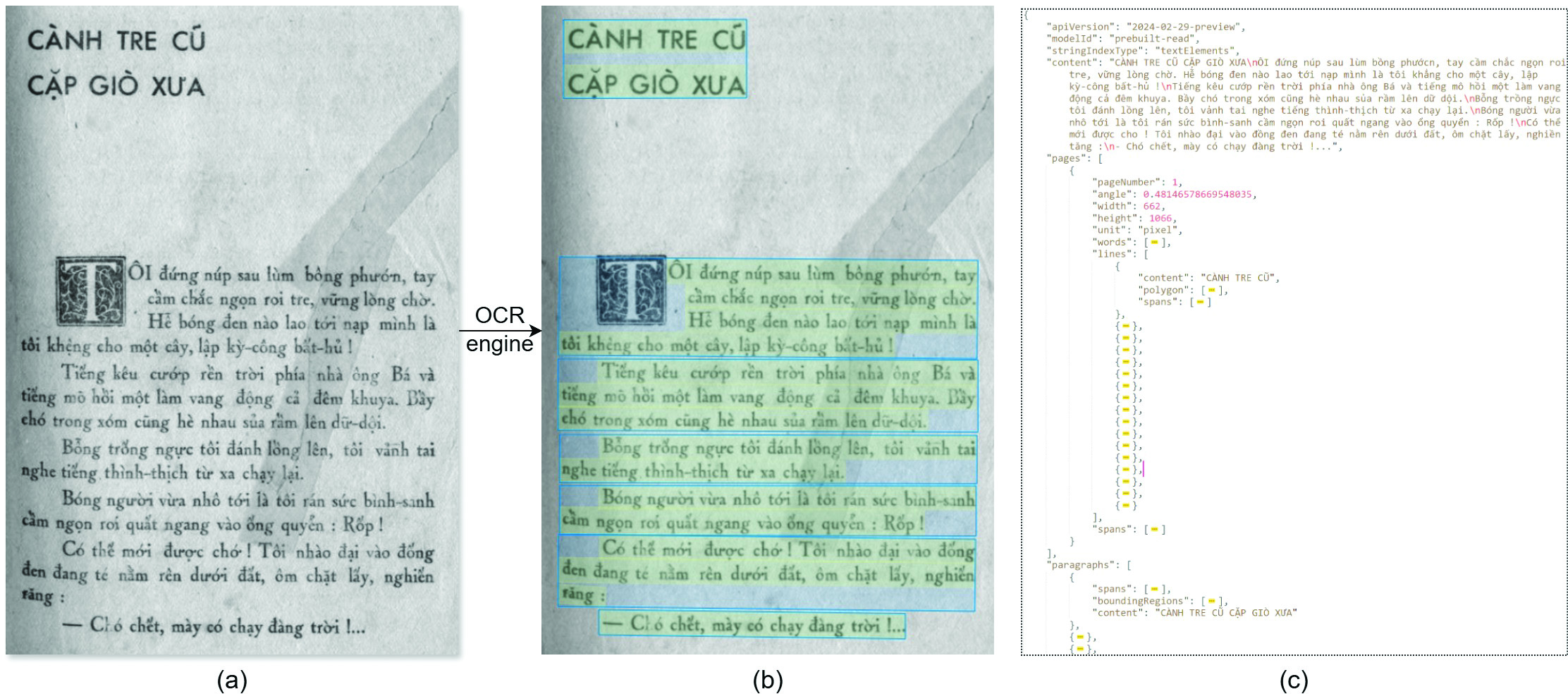}
    \caption{A sample OCR extraction by Azure DI:\\(a) the original page, (b) page visualized with paragraphs,\\(c) OCR extracted text in JSON form}
    \label{fig:azure_scheme}
\end{figure}

\subsection{Heuristic Filters to Reduce Noise}
Although ADI works well with plain text, it often produces noise when dealing with documents containing special objects such as watermarks, stamps, or background artifacts, which might reduce the quality of pseudo label eventually.

\begin{figure}[ht]
    \centering
    \includegraphics[width=0.47\textwidth]{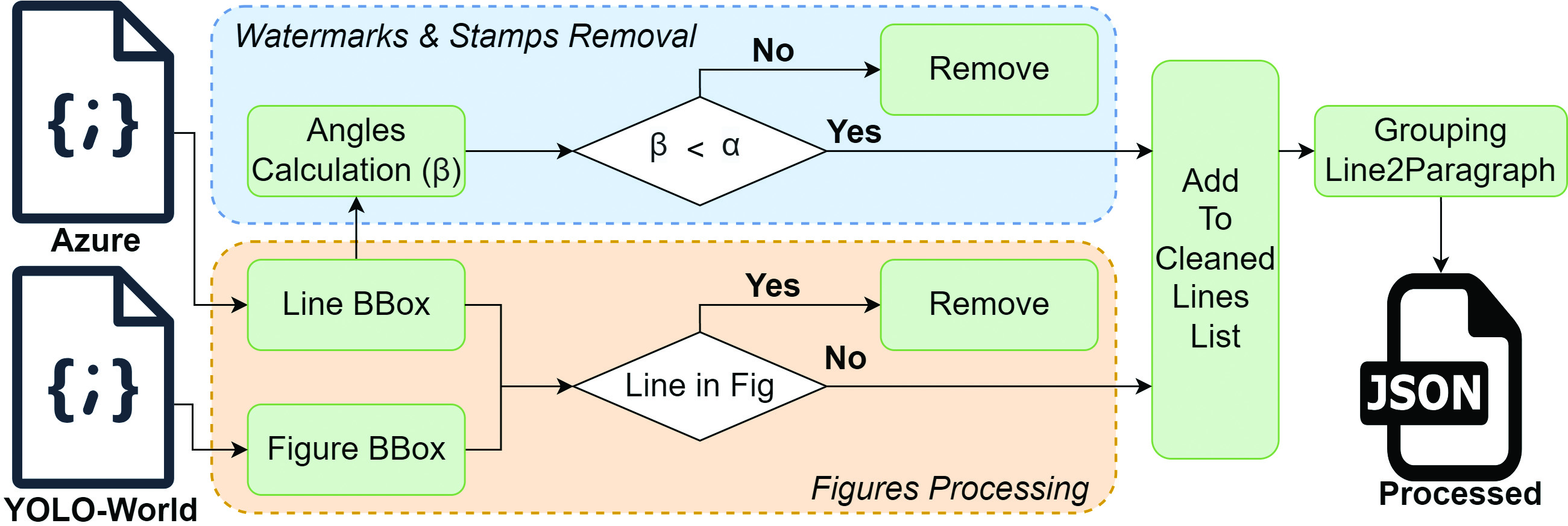}
    \caption{Noise Suppression by Heuristic Filters}
    \label{fig:json_proc}
\end{figure}

Based on the empirical observations, we propose heuristic filters to minimize noise. The procedure for cleaning the ADI output to make the \textbf{noise-suppressed} text is explained in Figure \ref{fig:json_proc}. The process consists of 2 sub-processes: \textit{Watermarks \& Stamps Removal} and \textit{Figures Processing}. 
We observe that most text within watermarks and stamps tends to be skewed with a large tilted angle of inclination relative to the horizontal. These cases are removed from the Watermarks \& Stamps Removal. A small amount of text within those noise that are not skewed or with a slight tilted angle is handled as a figure in the Figures Processing.

\begin{figure}[ht]
    \centering
    \includegraphics[width=0.4\textwidth]{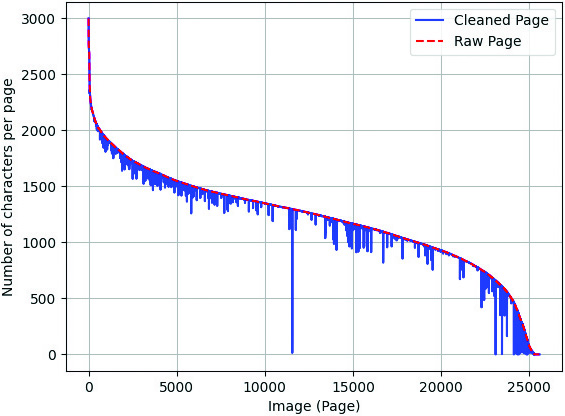}
    \caption{Number of characters per page between raw OCR text vs post-filtered text, sorted by raw character size}
    \label{fig:filter_effect}
\end{figure}

For \textit{Watermarks \& Stamps Removal}, we utilize the line's polygon information to compute the angle of inclination of the bounding box relative to the horizontal ($\beta$). We set an angle threshold ($\alpha$) to remove the skewed texts, which are the noise data. For \textit{Figures Processing}, we remove short text inside figures and some watermarks and stamps as we mentioned above by a simple rule: remove text within the bounding box of figures detected by the procedure described in the next subsection since these in-figure texts are irrelevant and not included in the ebook content. After these 2 subprocesses, we retain valid lines while the noise lines are ignored. Since our goal is to extract text at the paragraph level, we perform another step to group lines into paragraphs by using offset information provided by ADI. At this point, we have completed minimizing noise in the output of ADI. Figure \ref{fig:filter_effect} visualizes the effect of our heuristic filters in reducing the number of characters.

\subsection{Figure Detection by Bagging and Open-Vocab Objection Detection}
We used 2 vision LLMs: mPlug-DocOwl, LLaVa 1.6 to perform a zero-shot classification and one document analysis model LayoutParser \cite{shen2021layoutparser} to determine the existence of considerable figure(s) in the document image. Bagging is used here to filter out negative samples and reduce the number of likely positive samples that can be input into the Open-Vocab Object Detector (YOLO-World \cite{Cheng2024YOLOWorld}) for fine-grained detection by some proper prompt engineering. This arrangement performed well on the image sets with a 98.06\% F1 classification rate (manually verified).

\begin{figure}[ht]
    \centering
    \includegraphics[width=0.474\textwidth]{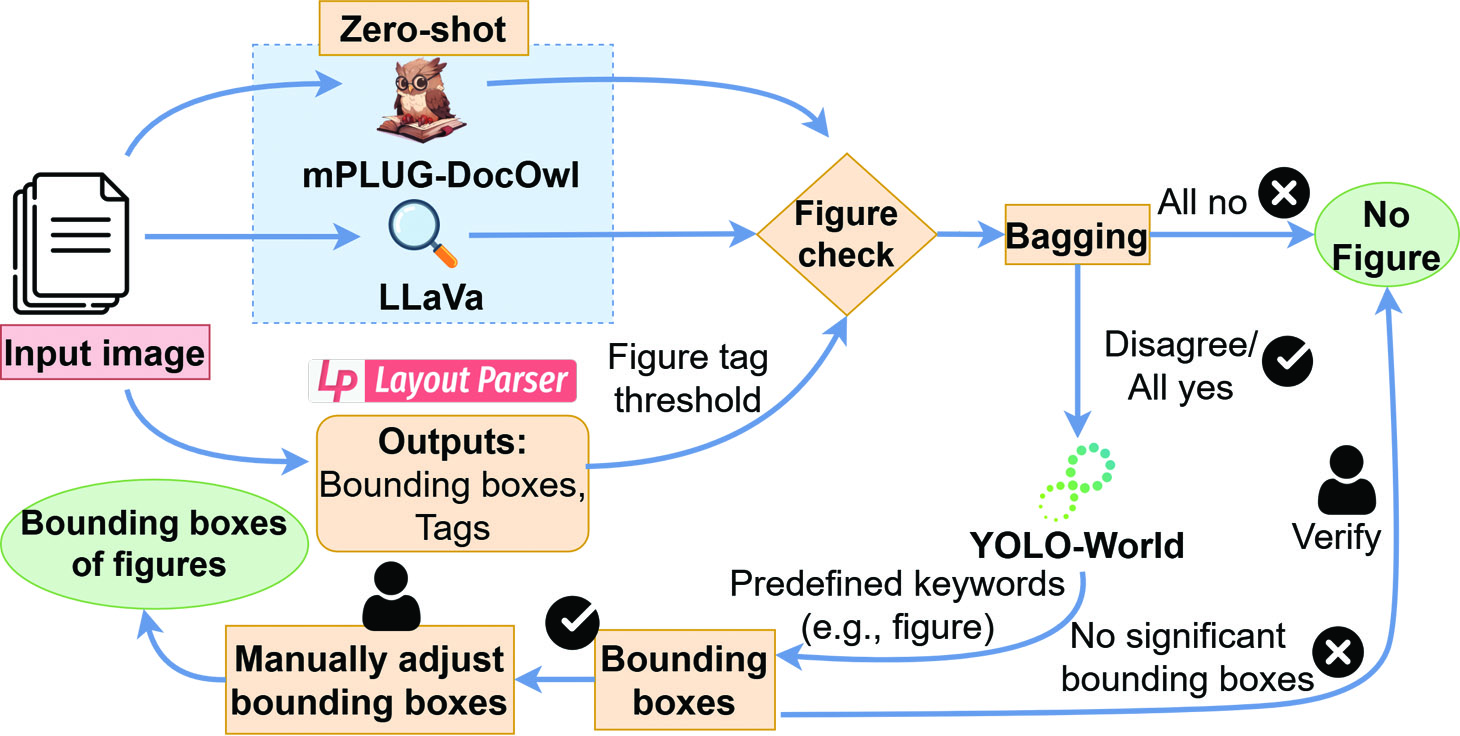}
    % \label{fig:json_processing}
    \caption{Figure Detection Procedure in Document}
    \label{fig:fig_det}
\end{figure}

The process is illustrated in Figure \ref{fig:fig_det} with a zero-shot prompt \textit{is there figure in the input document image?} Other parameters such as trivial size threshold, NMS threshold, and confident threshold are also tuned to detect good results. So far, to the best of our knowledge, this is a new zero-shot way to detect figures in document images.

\subsection{Localize Reference Text Region for Generated Paragraphs} \label{search_text}
Before clarifying this procedure, it is noteworthy to know that scanned pages occasionally contain stranded content (dangling text) where ebook creators have either neglected to transcribe the text or intentionally skipped it. This is because of various reasons such as explicit content, sensitivities, or censorship demands. Consequently, certain text segments extracted via OCR may not correspond to any content within the associated ebook.

Hence, generated texts at the paragraph level are simultaneously searched within the corresponding ebook (an HTML set) utilizing a fuzzy string matching based on Levenshtein distance to identify candidate textual regions that locally encapsulate the generated text. A similarity threshold is decided to consider if a match is \textit{found} or \textit{not found} by dangling content. In detail, textual windows of predefined block size from the HTML set are pooled and searched by computing similarity with the query (i.e. the text from OCR) using fuzzy matching. Found candidates are the windows having similarity greater than the threshold, while the best candidate that has the highest score is padded to ensure that the local reference covers enough contextual info for correction. This localization is designed to lower the cost and overload of the reference for LLMs to reduce issues such as hallucination or lengthy context. Later, paragraphs having perfect matches are left intact, whereas those with imperfect similarity are processed by an LLM given the best local reference found, as described in the next subsection. Short-length and dangling texts are processed in subsection \ref{post-matching}.

\subsection{Text Correction using LLM with Localize Reference Text Region}
From this step, we use Prompt 1 template to query an LLM for each text (paragraph) and obtain an answer before some hallucination mitigation.

{
\renewcommand{\tablename}{Prompt}
\setcounter{table}{0}
\begin{table}[h]
    \setlength{\abovecaptionskip}{0pt}
    \setlength{\belowcaptionskip}{0pt}
    \caption{OCR Correction by LLM with reference}
    \small % or \footnotesize, \small, etc., to shrink text
    \begin{tabular}{|p{0.93\columnwidth}|}
        \hline
        % Your prompt or query goes here; it will wrap automatically.
        Given an html content as below:
        
        \textbf{\{reference\_html}\} \newline
        
        Providing a text which may need correction as it is generated from an imperfect OCR tool.
        Use html content as reference source to correct below input text if needed:
        
        $<$input-text$>$\textbf{\{paragraph\}}$<$/input-text$>$ \newline
        
        Follow orthography, diacritic and punctuation of the reference, only correct words existed within input text!
        (other instructions for less hallucinated and well-formed answer)
        \\
        \hline
    \end{tabular}
    \label{tab:prompt1}
\end{table}
}

\textbf{Hallucination Occurrence:} LLMs have been found to hallucinate or generate text that was not part of the original input. This often occurs when the model tries to produce the most contextually probable continuation, which can lead to unintended completions when the input is cropped or incomplete \cite{lin2024towards}. Even with prompt engineering (explicit instructions), this behavior still happens at a rare rate for our task. For instance, let the input be \textit{ ``qulck bruwn fox jnnps''} while the reference piece is \textit{``Head padding. The quick brown fox jumps over the lazy dog. Tail padding''}, then LLM might (less likely) return such an answer like this \textit{``The quick brown fox jumps over the lazy dog.''} instead of the expected sequence \textit{``quick brown fox jumps''}.
Another rare case is processed texts including XML tags although there is a prevention directive in other instructions.

\textbf{Hallucination Mitigation:} For the XML tag inclusion issue, we carefully detect them by regular expression and unwrap the inner text. For the over-generation (unintended completions) problem, we propose a simple algorithm to take a suitable cropped window from the LLM answer to trim away words that weren't part of the original input. The simple algorithm does the following:
\begin{itemize}
    \item Split both original and LLM's answer texts into words.
    \item Slide a window of the same length $\pm 1$ at word-level as the original over the LLM answer.
    \item Calculate the similarity ratio by Levenshtein distance for each window.
    \item Choose the best matching window which the candidate has highest similarity.
\end{itemize}
In this way, we finally trim away hallucination parts from the head and tail of the answer to collect the best cropped of corrected text from the LLM.

\subsection{Minor Spelling Correction Given Context} \label{post-matching}
As we mentioned in subsection \ref{search_text} Reference Text Region Localization, there are texts that are short-length or not found anywhere because the ebook creators ignored them. These cases have finally undergone a minor spelling correction by the LLM if needed given its previous and following texts \textit{(if available)} on the same page. Below is the Prompt 2 template for this LLM-based spell correction:

{
\renewcommand{\tablename}{Prompt}
\setcounter{table}{1}
\begin{table}[h]
    \setlength{\abovecaptionskip}{0pt}
    \setlength{\belowcaptionskip}{0pt}
    \caption{Minor Spell Correction by LLM}
    \small % or \footnotesize, \small, etc., to shrink text
    \begin{tabular}{|p{0.93\columnwidth}|}
        \hline
        Given a text as below:\\
        $<$input-text$>$\textbf{\{text\}}$<$/input-text$>$ \newline
        
        Provide adjacent content nearby the input:
        $<$content$>$\textbf{\{nearby\}}$<$/content$>$ \newline
        
        Correct spelling only if needed (just minimal change!)
        Return the original if it is already fine (or it is random nonsense)
        (other instructions for well-formed answer)
        \\
        \hline
    \end{tabular}
    \label{tab:prompt2}
\end{table}
}

To mitigate hallucination in this step, we enforce a strict policy: LLM's answer must have high similarity as we want minimal change, a tiny difference in text lengths, and no change in the number of words. If the LLM returning fails these conditions, we reject the correction and keep the dangling text as original.

\section{Experiment}
In implementation, we chose OpenAI GPT-4o mini (2024-07-18) as the LLM for all the relevant tasks of our pipeline to build the dataset, as it is the most cost-efficient model having understanding in non-English languages such as Vietnamese, French. We used their batch API at reduced price with a 0 temperature for more consistent answers and minimal cost for the project. Other details of parameters used in our pipeline are described in the Appendix.

\subsection{Dataset Quality Evaluation}
To evaluate our Post-OCR processing, we set up 2 assessments: a human-critic evaluation on a subset, and an automatic evaluation for the whole dataset.

Due to the large volume, we manually evaluated only a stratified random subset. Specifically, we sampled 1025 instances from 123 books, which were then reviewed by human evaluators who had received and understood properly the guidelines. Vietnamese evaluators were asked to see the page, and grade transcriptions from 1 to 10 from no-match to perfect respectively. A perfect score (10) requires up to 1 subtle word-level error. From scores 9 to 6, every 2 errors reduce the score by 1, and from scores 5 to 2, every 3 errors reduce it by 1. Due to availability, only 450 samples were ultimately evaluated. 
We averaged the scores given by evaluators and compared ours with \cite{tran2021hierarchical} approach which publicly available. Our approach scored a mean of 8.72 with a median of 8.8, SD$=0.63$ while \cite{tran2021hierarchical} obtained a mean of 7.03 with a median of 7.46, SD$=0.94$.

\setcounter{table}{1}
\begin{table*}[t]
\centering
\begin{tabular}{lcccccccc}
\hline
\textbf{Method} & \textbf{Norm Edit Dist $\downarrow$} & \textbf{BLEU $\uparrow$} & \textbf{Precision $\uparrow$} & \textbf{Recall $\uparrow$} & \textbf{F1 $\uparrow$} & \textbf{CER $\downarrow$} & \textbf{WER $\downarrow$} \\ \hline
GPT-4o    & \underline{0.05} &\textbf{ 0.81} & \textbf{0.90} & \textbf{0.91} & \textbf{0.90} & \underline{0.05} & \textbf{0.11} \\
Azure Doc AI & \textbf{0.04} & \underline{0.76} & \underline{0.89} & \underline{0.87} & \underline{0.88} & \textbf{0.04} & \underline{0.13} \\
Google Doc AI & 0.07 & 0.75 & \underline{0.89} & \underline{0.87} & \underline{0.88} & 0.07 & 0.14 \\ \hline
Tesseract & \underline{0.12} & \underline{0.60} & \underline{0.79} & \underline{0.75} & \underline{0.76} & \underline{0.14} & \underline{0.27} \\
EasyOCR & 0.24 & 0.20 & 0.52 & 0.45 & 0.48 & 0.25 & 0.59 \\ 
Donut Finetuned (Ours)      & 0.14 & 0.28 & 0.58 & 0.58 & 0.58 & 0.15 & 0.44 \\
Tesseract Finetuned (Ours)      & \textbf{0.08} & \textbf{0.71} & \textbf{0.86} & \textbf{0.84} & \textbf{0.85} & \textbf{0.09} & \textbf{0.18} \\ \hline
\end{tabular}
\caption{Result comparison between our baselines and other methods: GPT-4o, Azure DI, Google Doc AI, Tesseract and EasyOCR. The upper group is for closed-source models while the lower group is for open-source methods. Bold, underlined numbers are the \textbf{best}, and \underline{second-best} results respectively for each group.}
\label{tab:result_comparison}
\end{table*}

An automatic way was also proposed to evaluate relatively the whole set: we prompt GPT-4o to compare carefully 2 versions of texts by ours and \cite{tran2021hierarchical} spell-correction with explanation; with criteria is that the better means the transcript has more sense (clearer semantics), regardless the language dialect. The result is 86.66\% ours is better, only 6.81\% \cite{tran2021hierarchical} result is better while 6.5\% are equal. These assessments suggest our method yields a better pseudo label compared to the transformer-based spelling correction. Note that \cite{tran2021hierarchical} method was input by our noise-suppressed texts for a fair comparison.

\subsection{Baseline Models}
We employ the Donut \cite{kim2022ocr}, as our base model for full-page recognition. This architecture consists of a vision encoder Swin Transformer \cite{liu2021swin} that converts the image to an embedding and a text decoder BART \cite{lewis2019bart} that generates text based on the embedding. We use the pretrained Donut model, to fine-tune on our dataset. The fine-tuning process focused on reading all texts in the image from top left to bottom right, with the objective of minimizing cross-entropy loss of the next token prediction by considering both the image and previous contexts. In addition, we employ Tesseract OCR \cite{smith2007overview} version 5 as the second baseline of the 2-stage engine. Using our dataset but at line level, we finetuned the recognition module from the best Vietnamese pretrained.

\textbf{Setups} We split our dataset into train, val, test, with ratios of 8:1:1 respectively. For \textit{Donut baseline}, we resize the pages to $960 \times 1280$ pixels, with a max length of 1536 in the decoder. The batch size is 16, and 30 epochs for each experiment. The initial learning rate is set to $1e-4$ with cosine scheduling. One experiment takes around 3 days on 4 A100 40GB GPUs. For \textit{Tesseract baseline}, we resize the line image (cropped by polygon) to half of its original resolution before inputting it into training at a batch size of 1, a learning rate of $5e-5$ and 500,000 iterations. It takes around 24 hours for each experiment.

To support the \textit{Donut baseline}, a synthetic set is created by sampling blank pages with various noise backgrounds. We used a GPT-2 \cite{radford2019language} based model fine-tuned on a Vietnamese corpus to generate long texts in various topics. These texts were synthesized to the blank pages with varied layouts and styles to create 100,000 samples.

\subsection{OCR Evaluation}
We use common OCR task metrics to evaluate: \textbf{Norm Edit Distance} (normalized character-level discrepancies); \textbf{F1} score (harmonic mean of precision and recall); \textbf{BLEU} (a metric from machine translation, applied to OCR \cite{papineni2002bleu}); \textbf{CER} (Character Error Rate, percentage of incorrect characters); and \textbf{WER} (Word Error Rate, percentage of incorrect words).

We compare our baselines with commonly used models including the multi-modal GPT-4o \cite{openai2024gpt4o}, Tesseract, Azure DI, Google Document AI and EasyOCR. We asked GPT-4o to perform OCR on page images at temperature of 0, max token limit of 4096. Images were first resized to $960 \times 1280$ before the task was executed using their batch API. It took approximately 8 hours to complete. Similarly, Azure DI and Google Doc AI returned results within 7 and 14 hours respectively due to their query limitations. 

The result is reported in Table ~\ref{tab:result_comparison}. Closed-source models like GPT-4o, Azure DI, and Google Doc AI show great results with GPT-4o achieving the best in almost all metrics. Among open-source methods, our Tesseract finetuned on VieBookRead dataset outperforms the others, achieving decent results quite close to the metrics of Google Doc AI. Notably, after fine-tuning, Tesseract's edit distance decreased from 0.12 to just 0.08, and sharply from 0.27 to 0.18 on word-level error. This demonstrates the usefulness of our dataset for the task. Besides, Donut fine-tuned on our dataset performs moderately suggesting that the full-page recognition model needs to be improved compared with the 2-stage approach like Tesseract. In general, VieBookRead is useful for training the OCR engine in Vietnamese and could make a significant contribution to OCR development.

\section{Conclusion}
From our experiments, we found that dense text documents, such as books, remain challenging for unified models like Donut while 2-stage models like Tesseract perform better. We hypothesize that this is due to the recognition part of 2-stage models handling only smaller regions, hence lower model complexity and hallucination. Still, 2-stage models may struggle with degraded text, where contextual info from nearby lines assists correct prediction actually.

In conclusion, we present a method to use ebook resources and LLM to correct post-OCR text, generating fine-grained pseudo ground truth without annotating costs, while preserving the historical characteristics of classical documents. This approach applies to Vietnamese and other diacritical languages, facilitating the creation of a quality page2page dataset to improve open-source OCR models for underserved communities. Finally, we introduce VieBookRead, the first large publicly available Vietnamese dataset for 19th-century classical books, to bridge gaps in existing datasets and support further research.

\section*{Acknowledgments}
This work was supported by the Innovative Human Resource Development for Local Intellectualization program through  IITP (Institute of Information \& Communications Technology Planning \& Evaluation) grant (IITP-2024-RS-2020-II201489, 50\%) and the IITP-ITRC (Information Technology Research Center) (IITP-2024-RS-2023-00259703, 50\%) grant funded by the Korea government (MSIT).

The work was also supported by Hyundai Motor Chung Mong-Koo Global Scholarship to Thao Do (1st author) and An Vo (3rd author). We are also thankful for the API research credits from OpenAI to An Vo and Google Cloud Platform to Thao Do.

\bibliography{aaai25}

\begin{thebibliography}{26}
\providecommand{\natexlab}[1]{#1}

\bibitem[{Atserias et~al.(2012)Atserias, Fort, Nazar, and Renau}]{atserias2012spell}
Atserias, J.; Fort, M.~F.; Nazar, R.; and Renau, I. 2012.
\newblock Spell Checking in Spanish: The Case of Diacritic Accents.
\newblock In \emph{LREC}, 737--742.

\bibitem[{Cheng et~al.(2024)Cheng, Song, Ge, Liu, Wang, and Shan}]{Cheng2024YOLOWorld}
Cheng, T.; Song, L.; Ge, Y.; Liu, W.; Wang, X.; and Shan, Y. 2024.
\newblock YOLO-World: Real-Time Open-Vocabulary Object Detection.
\newblock In \emph{Proc. IEEE Conf. Computer Vision and Pattern Recognition (CVPR)}.

\bibitem[{Dang et~al.(2022)Dang, Nguyen, Pham, Nguyen, Chau, Ngo, Nguyen, Phan, Trinh, Nguyen et~al.}]{dang2022nomnaocr}
Dang, H.-Q.; Nguyen, D.-A.; Pham, P.-P.; Nguyen, N.-T.; Chau, T.; Ngo, D.-V.; Nguyen, T.-H.; Phan, C.-T.; Trinh, T.-H.; Nguyen, M.-T.; et~al. 2022.
\newblock Nomnaocr: The first dataset for optical character recognition on han-nom script.
\newblock In \emph{2022 RIVF International Conference on Computing and Communication Technologies (RIVF)}, 476--481. IEEE.

\bibitem[{Do et~al.(2021)Do, Nguyen, Bui, and Vo}]{do2021vsec}
Do, D.-T.; Nguyen, H.~T.; Bui, T.~N.; and Vo, H.~D. 2021.
\newblock Vsec: Transformer-based model for vietnamese spelling correction.
\newblock In \emph{PRICAI 2021: Trends in Artificial Intelligence: 18th Pacific Rim International Conference on Artificial Intelligence, PRICAI 2021, Hanoi, Vietnam, November 8--12, 2021, Proceedings, Part II 18}, 259--272. Springer.

\bibitem[{Gooding, Terras, and Warwick(2013)}]{gooding2013myth}
Gooding, P.; Terras, M.; and Warwick, C. 2013.
\newblock The myth of the new: mass digitization, distant reading, and the future of the book.
\newblock \emph{Literary and Linguistic Computing}, 28(4): 629--639.

\bibitem[{Hl{\'a}dek, Sta{\v{s}}, and Pleva(2020)}]{hladek2020survey}
Hl{\'a}dek, D.; Sta{\v{s}}, J.; and Pleva, M. 2020.
\newblock Survey of automatic spelling correction.
\newblock \emph{Electronics}, 9(10): 1670.

\bibitem[{Kim et~al.(2022)Kim, Hong, Yim, Nam, Park, Yim, Hwang, Yun, Han, and Park}]{kim2022ocr}
Kim, G.; Hong, T.; Yim, M.; Nam, J.; Park, J.; Yim, J.; Hwang, W.; Yun, S.; Han, D.; and Park, S. 2022.
\newblock Ocr-free document understanding transformer.
\newblock In \emph{European Conference on Computer Vision}, 498--517. Springer.

\bibitem[{Le, Mai, and Lam(2023)}]{le2023dataset}
Le, A.; Mai, D. T.~H.; and Lam, T. 2023.
\newblock A Dataset of Vietnamese Documents for Text Detection.
\newblock In \emph{International Conference on Future Data and Security Engineering}, 418--429. Springer.

\bibitem[{Lewis et~al.(2019)Lewis, Liu, Goyal, Ghazvininejad, Mohamed, Levy, Stoyanov, and Zettlemoyer}]{lewis2019bart}
Lewis, M.; Liu, Y.; Goyal, N.; Ghazvininejad, M.; Mohamed, A.; Levy, O.; Stoyanov, V.; and Zettlemoyer, L. 2019.
\newblock Bart: Denoising sequence-to-sequence pre-training for natural language generation, translation, and comprehension.
\newblock \emph{arXiv preprint arXiv:1910.13461}.

\bibitem[{Lin et~al.(2024)Lin, Guan, Zhang, Zhang, Li, and Zhang}]{lin2024towards}
Lin, Z.; Guan, S.; Zhang, W.; Zhang, H.; Li, Y.; and Zhang, H. 2024.
\newblock Towards trustworthy LLMs: a review on debiasing and dehallucinating in large language models.
\newblock \emph{Artificial Intelligence Review}, 57(9): 1--50.

\bibitem[{Liu et~al.(2021)Liu, Lin, Cao, Hu, Wei, Zhang, Lin, and Guo}]{liu2021swin}
Liu, Z.; Lin, Y.; Cao, Y.; Hu, H.; Wei, Y.; Zhang, Z.; Lin, S.; and Guo, B. 2021.
\newblock Swin transformer: Hierarchical vision transformer using shifted windows.
\newblock In \emph{Proceedings of the IEEE/CVF international conference on computer vision}, 10012--10022.

\bibitem[{Nguyen, Dang, and Nguyen(2020)}]{nguyen2020deep}
Nguyen, H.~T.; Dang, T.~B.; and Nguyen, L.~M. 2020.
\newblock Deep learning approach for vietnamese consonant misspell correction.
\newblock In \emph{Computational Linguistics: 16th International Conference of the Pacific Association for Computational Linguistics, PACLING 2019, Hanoi, Vietnam, October 11--13, 2019, Revised Selected Papers 16}, 497--504. Springer.

\bibitem[{Nguyen et~al.(2008)Nguyen, Ngo, Phan, Dinh, and Huynh}]{nguyen2008vietnamese}
Nguyen, P.~H.; Ngo, T.~D.; Phan, D.~A.; Dinh, T.~P.; and Huynh, T.~Q. 2008.
\newblock Vietnamese spelling detection and correction using Bi-gram, Minimum Edit Distance, SoundEx algorithms with some additional heuristics.
\newblock In \emph{2008 IEEE International Conference on Research, Innovation and Vision for the Future in Computing and Communication Technologies}, 96--102. IEEE.

\bibitem[{OpenAI(2024)}]{openai2024gpt4o}
OpenAI. 2024.
\newblock Hello GPT-4o.
\newblock \url{https://openai.com/index/hello-gpt-4o/}.

\bibitem[{Papineni et~al.(2002)Papineni, Roukos, Ward, and Zhu}]{papineni2002bleu}
Papineni, K.; Roukos, S.; Ward, T.; and Zhu, W.-J. 2002.
\newblock Bleu: a method for automatic evaluation of machine translation.
\newblock In \emph{Proceedings of the 40th annual meeting of the Association for Computational Linguistics}, 311--318.

\bibitem[{Radford et~al.(2021)Radford, Kim, Hallacy, Ramesh, Goh, Agarwal, Sastry, Askell, Mishkin, Clark et~al.}]{radford2021learning}
Radford, A.; Kim, J.~W.; Hallacy, C.; Ramesh, A.; Goh, G.; Agarwal, S.; Sastry, G.; Askell, A.; Mishkin, P.; Clark, J.; et~al. 2021.
\newblock Learning transferable visual models from natural language supervision.
\newblock In \emph{International conference on machine learning}, 8748--8763. PMLR.

\bibitem[{Radford et~al.(2019)Radford, Wu, Child, Luan, Amodei, Sutskever et~al.}]{radford2019language}
Radford, A.; Wu, J.; Child, R.; Luan, D.; Amodei, D.; Sutskever, I.; et~al. 2019.
\newblock Language models are unsupervised multitask learners.
\newblock \emph{OpenAI blog}, 1(8): 9.

\bibitem[{Ruch, Baud, and Geissb{\"u}hler(2003)}]{ruch2003using}
Ruch, P.; Baud, R.; and Geissb{\"u}hler, A. 2003.
\newblock Using lexical disambiguation and named-entity recognition to improve spelling correction in the electronic patient record.
\newblock \emph{Artificial intelligence in medicine}, 29(1-2): 169--184.

\bibitem[{Shen et~al.(2021)Shen, Zhang, Dell, Lee, Carlson, and Li}]{shen2021layoutparser}
Shen, Z.; Zhang, R.; Dell, M.; Lee, B. C.~G.; Carlson, J.; and Li, W. 2021.
\newblock LayoutParser: A Unified Toolkit for Deep Learning Based Document Image Analysis.
\newblock \emph{arXiv preprint arXiv:2103.15348}.

\bibitem[{Smith(2007)}]{smith2007overview}
Smith, R. 2007.
\newblock An overview of the Tesseract OCR engine.
\newblock In \emph{Ninth international conference on document analysis and recognition (ICDAR 2007)}, volume~2, 629--633. IEEE.

\bibitem[{Stankevi{\v{c}}ius et~al.(2022)Stankevi{\v{c}}ius, Luko{\v{s}}evi{\v{c}}ius, Kapo{\v{c}}i{\=u}t{\.e}-Dzikien{\.e}, Briedien{\.e}, and Krilavi{\v{c}}ius}]{stankevivcius2022correcting}
Stankevi{\v{c}}ius, L.; Luko{\v{s}}evi{\v{c}}ius, M.; Kapo{\v{c}}i{\=u}t{\.e}-Dzikien{\.e}, J.; Briedien{\.e}, M.; and Krilavi{\v{c}}ius, T. 2022.
\newblock Correcting diacritics and typos with a ByT5 transformer model.
\newblock \emph{Applied Sciences}, 12(5): 2636.

\bibitem[{Thi et~al.(2022)Thi, Vo, Nguyen, Nguyen, Tran, Dang, Quach, and Do}]{thi2022novel}
Thi, T.~N.; Vo, K.-H.; Nguyen, M.-T.; Nguyen, T.-T.; Tran, G.-P.~P.; Dang, C.-T.; Quach, C.-T.; and Do, T.-H. 2022.
\newblock A novel two-stages information extraction algorithm for vietnamese book cover images.
\newblock In \emph{International Symposium on Intelligent and Distributed Computing}, 160--167. Springer.

\bibitem[{Thomas, Gaizauskas, and Lu(2024)}]{thomas2024leveraging}
Thomas, A.; Gaizauskas, R.; and Lu, H. 2024.
\newblock Leveraging LLMs for Post-OCR Correction of Historical Newspapers.
\newblock In \emph{Proceedings of the Third Workshop on Language Technologies for Historical and Ancient Languages (LT4HALA)@ LREC-COLING-2024}, 116--121.

\bibitem[{Touvron et~al.(2023)Touvron, Martin, Stone, Albert, Almahairi, Babaei, Bashlykov, Batra, Bhargava, Bhosale et~al.}]{touvron2023llama}
Touvron, H.; Martin, L.; Stone, K.; Albert, P.; Almahairi, A.; Babaei, Y.; Bashlykov, N.; Batra, S.; Bhargava, P.; Bhosale, S.; et~al. 2023.
\newblock Llama 2: Open foundation and fine-tuned chat models.
\newblock \emph{arXiv preprint arXiv:2307.09288}.

\bibitem[{Tran et~al.(2021)Tran, Dinh, Phan, and Nguyen}]{tran2021hierarchical}
Tran, H.; Dinh, C.~V.; Phan, L.; and Nguyen, S.~T. 2021.
\newblock Hierarchical transformer encoders for Vietnamese spelling correction.
\newblock In \emph{Advances and Trends in Artificial Intelligence. Artificial Intelligence Practices: 34th International Conference on Industrial, Engineering and Other Applications of Applied Intelligent Systems, IEA/AIE 2021, Kuala Lumpur, Malaysia, July 26--29, 2021, Proceedings, Part I 34}, 547--556. Springer.

\bibitem[{Van~Nguyen et~al.(2024)Van~Nguyen, Tran, Pham, Nguyen, Nguyen, Van~Nguyen, and Nguyen}]{van2024vitextvqa}
Van~Nguyen, Q.; Tran, D.~Q.; Pham, H.~Q.; Nguyen, T. K.-B.; Nguyen, N.~H.; Van~Nguyen, K.; and Nguyen, N. L.-T. 2024.
\newblock ViTextVQA: A Large-Scale Visual Question Answering Dataset for Evaluating Vietnamese Text Comprehension in Images.
\newblock \emph{arXiv preprint arXiv:2404.10652}.

\end{thebibliography}

\end{document}